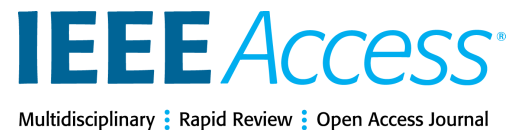

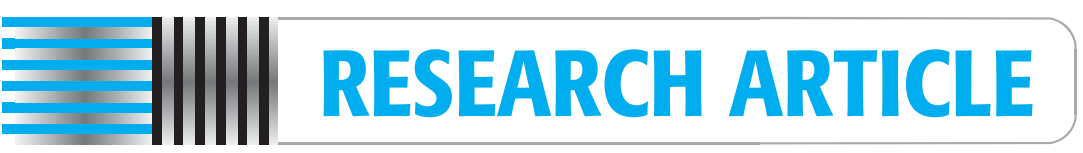

# Robust Semi-Supervised Regression for Vehicle Interior Noise Prediction

**SEJIN SIM, JINSOO BAE, AND SEOUNG BUM KIM**
School of Industrial and Management Engineering, Korea University, Seoul 02841, South Korea

Corresponding author: Seoung Bum Kim (sbkim1@korea.ac.kr)

This work was supported in part by the Brain Korea 21 FOUR, and in part by the Korea Institute for Advancement of Technology (KIAT) funded by the Korean Government (MOTIE) (The Competency Development Program for Industry Specialist) under Grant P0008691.

**ABSTRACT** The rapid advancement of artificial intelligence has observed increased application in predicting vehicle interior noise levels within the automotive industry. However, the collection of labeled data for training models in this context involves significant costs. Previous studies in semi-supervised regression (SSR) have effectively mitigated the reliance on labeled data by incorporating unlabeled data. Nonetheless, these approaches often introduce a high computational cost due to the training of multiple models and data sampling. This study introduces SpecRegMatch, a novel SSR method aimed at addressing the computational cost associated with training by leveraging a single model, thus eliminating the need for multiple data samplings. SpecRegMatch integrates consistency regularization and information maximization to robustly train the model, achieved through various augmentations applied to both the embedding vectors and predicted values. Experimental results demonstrate that SpecRegMatch achieves state-of-the-art performance across various scenarios, even when using a single model. It attains a remarkable performance, as indicated by an $R^2$ score of 0.434. This is especially noteworthy in scenarios where labeled data is scarce. You can access the code for our proposed method at https://github.com/sejin-sim/SpecRegMatch.

## I. INTRODUCTION

The advancement of artificial intelligence has considerably broadened its application, successfully addressing intricate challenges across diverse fields such as finance, healthcare, pharmaceuticals, and manufacturing [1], [2], [3], [4], [5], [6], [7]. In the automotive industry, research has harnessed deep learning techniques to meet the increasing consumer demand for quieter vehicle environments [8], [9], [10], [11], [12], [13]. In the development of automobile models, traditionally, evaluating the interior noise level of vehicles to ascertain vehicle quality necessitated direct measurements. Nevertheless, to mitigate the expense associated with direct measurements, studies have explored the prediction of noise levels in vehicles through the use of deep learning models. Recent studies aiming to predict vehicle interior noise levels have adopted spectrograms as the primary input data for deep learning models. Spectrograms, derived from signal data such as acceleration data, enable the model to extract comprehensive information [11], [12], [13], [14]. A spectrogram serves as a transformed representation of signal data, aiding visualization and analysis by facilitating the extraction of crucial features such as frequency bands [15], [16]. While numerous previous studies have concentrated on a single representative value for the entire frequency band [12], [13], it is crucial to acknowledge that human perception of the impact of interior noise varies across different frequency bands. This underscores the necessity for a deep learning model capable of predicting vehicle interior noise levels based on specific frequency bands, enabling a thorough analysis of vehicle quality [17].

Specialized experts are required to independently gather interior noise levels and acceleration data using dedicated equipment for training a deep learning model. Figure 1 illustrates the data collection process. In the driving phase, sensors affixed to the devices capture acceleration data, while a microphone positioned near the driver's seat collects vehicle interior noise levels across various frequency bands. The devices equipped with sensors for collecting signal data consist of a balt-nut assembly (BNA) and a motor, each providing six sets of acceleration data from their respective X, Y, and Z axes. These data collection processes require expensive equipment and the expertise of professionals to ensure accurate measurements.

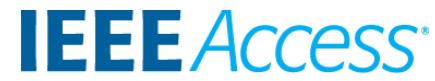

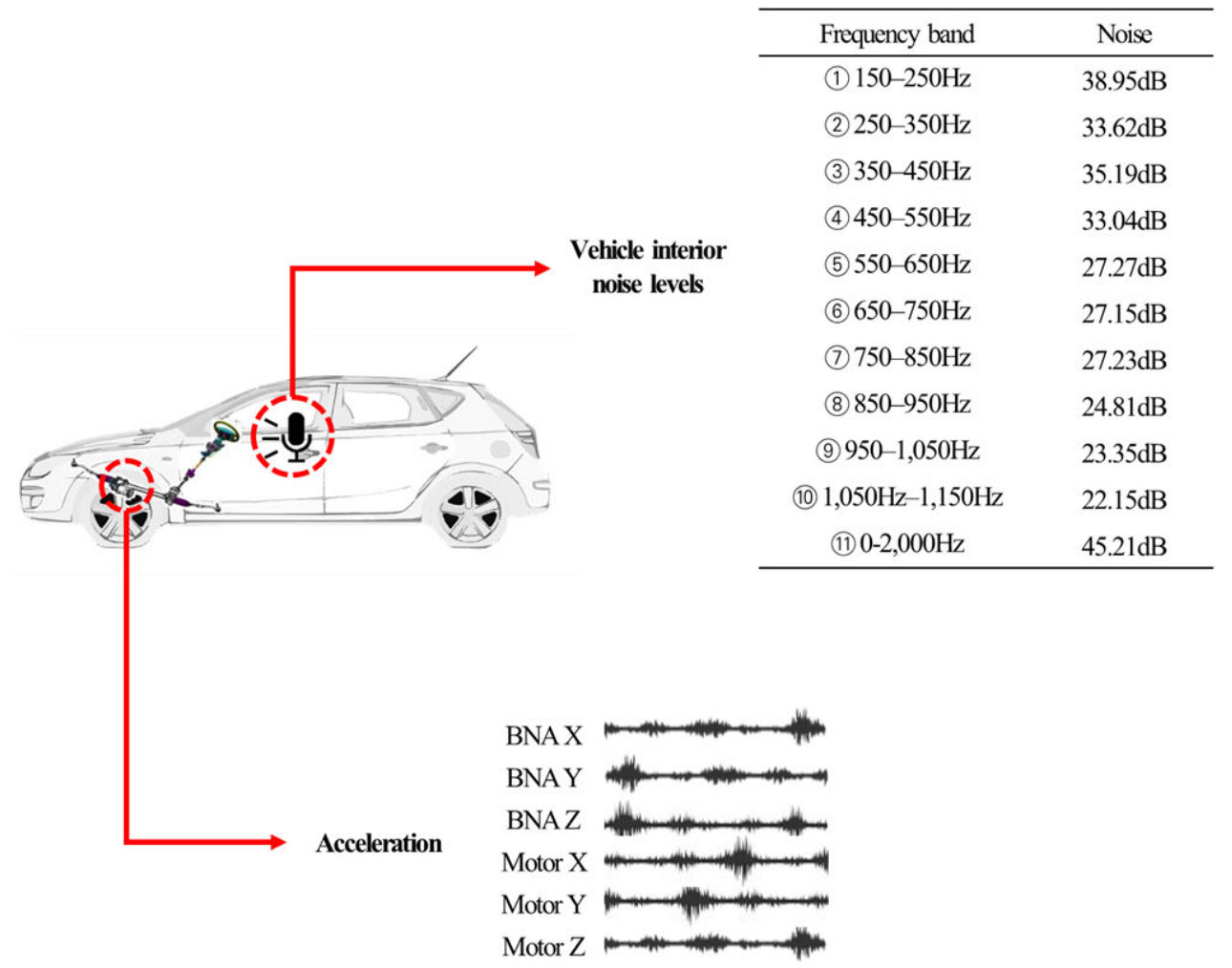

**FIGURE 1. Example of the data collection process.**

A field dataset comprising only acceleration data that can be easily generated in environments similar to a real environment is considered unlabeled data. In the automotive industry, a prior study used generative adversarial neural networks (GAN) to tackle the issue of high costs related to data collection by creating synthetic samples across diverse labeled data scenarios [18]. Nevertheless, the utilization of GANs for generating synthetic samples may give rise to mode collapse issues, leading to the production of predominantly similar data [19]. Therefore, opting for unlabeled data collected in an environment akin to labeled data can offer higher stability in model training. Semi-supervised learning (SSL) has received considerable attention as a means of overcoming the challenge of limited labeled data. SSL enables the deep learning models to leverage unlabeled data to obtain additional information, thereby enhancing their prediction abilities [20]. Previous studies on semi-supervised regression (SSR) involved training multiple models to generate high-quality pseudo-labels for unlabeled data [2], [3], [4], [5], [6], [7], [8], [9], [10], [11], [12], [13], [14], [15], [16], [17], [18], [19], [20], [21], [22], [23]. More recently, the ensemble Bayesian neural network (BNN) method was introduced to adjust for pseudo-label quality of unlabeled data by estimating uncertainty associated with unlabeled data [21]. However, a BNN requires multiple data samplings for posterior distribution estimation, and methods such as ensemble or co-training require a high computational cost during model training [24]. Therefore, SSR methods with a simple structure and single-model training are necessary to address these challenges.

This study proposes SpecRegMatch, a semi-supervised method that addresses the high computational cost of training using a single model in regression problems without requiring multiple data samplings. SpecRegMatch incorporates consistency regularization and information maximization to enhance the model's robustness. This is achieved by applying various augmentations to the spectrogram and using these augmented spectrograms for embedding vectors and predicted values.

For consistency regularization, SpecRegMatch uses predictions from the weakly augmented, strongly augmented, and mixup-augmented spectrograms, with the prediction of a weakly augmented spectrogram as a pseudo-label. Weak augmentation involves applying a minimal perturbation to the original spectrogram, while strong augmentation entails a more pronounced perturbation. Mixup augmentation combines weakly and strongly augmented spectrograms in a linear manner. For information maximization, SpecRegMatch computes the cross-correlation matrix from embedding vectors of the weakly and strongly augmented spectrograms and approximates it by training a model with the identity matrix. Information maximization can make the embedding vectors of differently augmented versions of a spectrogram more similar while minimizing the redundancy among the components of embedding vectors, ensuring that the embedding vector contains non-redundant information about the spectrogram.

Furthermore, SpecRegMatch is a multi-output structure capable of predicting vehicle interior noise levels across various frequency bands, allowing for an in-depth analysis of vehicle quality. This study conducts predictions for noise levels in ten individual frequency bands, ranging from 150 Hz to 1,150 Hz at intervals of 100 Hz, as well as the entire frequency band (0–2,000 Hz). Spectrograms derived from six devices are used as inputs to the prediction model, enabling accurate predictions of vehicle interior noise levels based on multiple frequency bands and extracting more comprehensive information from acceleration data. While spectrograms have been used in previous semi-supervised classification studies [25], [26], [27], their application within the context of regression (when the output is continuous) has yet to be explored. To the best of our knowledge, this study is the first to apply SSR analysis to spectrograms. The effectiveness of SpecRegMatch is evaluated and compared with other SSR methods.

The contributions of this study are summarized as follows:

1) This study is the first, to our knowledge, to use spectrograms for SSR, expanding its potential applications across various signal data scenarios.
2) The proposed SSR method incorporates consistency regularization through predictions of the weakly

**TABLE 1. Overview of representative semi-supervised learning method.**

| Category | Method | Description |
|---|---|---|
| Semi-supervised classification | MixMatch [28] | Averages predictions from unlabeled data with various augmentations and uses a sharpening technique to refine class probabilities |
| | ReMixMatch [29] | Integrates distribution alignment and augmentation anchoring into MixMatch's framework |
| | FixMatch [30] | Simplifies the combination of consistency regularization and pseudo-labeling with strong augmentation |
| | FlexMatch [31] | Adjusts the flexible thresholds for curriculum pseudo-labeling |
| | SimMatch [32] | Matches the similarity relationships of both semantic and instance levels using a labeled memory buffer. |
| Semi-supervised regression | COREG [22] | Uses two models to generate pseudo-labels from unlabeled data |
| | SSDKL [33] | Combines neural networks and kernel methods to minimize prediction variance of unlabeled data |
| | TNNR [23] | Trains the difference between predictions of two unlabeled data pairs through loop consistency |
| | UCVME [21] | Improves the quality of uncertainty estimates on unlabeled data using a Bayesian framework and model ensemble |

augmented, strongly augmented, and mixup-augmented spectrograms. Information maximization is achieved using embedding vectors to robustly train a model with various augmentations, even with limited label data.

3) In comparison to recently proposed SSR methods that require multiple models and involve multiple data samplings, SpecRegMatch utilizes a single model, reducing computational costs and offering an intuitive and straightforward approach.

The remainder of this paper is organized as follows: Section II reviews previous studies related to the SSR method. Section III introduces the proposed SpecRegMatch architecture and methods. Section IV describes the experimental setup and discusses the results obtained using the vehicle interior noise dataset. Finally, Section V concludes the paper and discusses future research directions.

## II. RELATED WORKS

### A. SEMI-SUPERVISED LEARNING METHODS

SSL reduces reliance on labeled data using unlabeled data. Recently, hybrid methods combining various techniques, such as consistency regularization, pseudo-labeling, and data augmentation, have gained attention in SSL [20]. Table 1 provides an overview of representative SSL methods for classification and regression problems.

#### 1) SEMI-SUPERVISED CLASSIFICATION

Berthelot et al. [28] proposed MixMatch, which averages predictions from unlabeled data with various augmentations and uses a sharpening technique to reduce the entropy of class probabilities for unlabeled data. Berthelot et al. [29] proposed ReMixMatch, an extension of MixMatch that incorporates distribution alignment and augmentation anchoring. ReMixMatch uses distributional alignment to make the distribution of predictions on unlabeled data similar to that of labeled data. Additionally, it introduces augmentation anchoring that uses the prediction for weakly augmented data as the target for predictions on strongly augmented data. However, these approaches entail high computational costs because of the integration of various techniques. To overcome these limitations, Sohn et al. [30] proposed FixMatch, a simplified approach combining consistency regularization and pseudo-labeling to mitigate the high computational costs of previous methods. FixMatch uses a consistency regularization technique that minimizes the difference between the predictions of weak and strong augmentations, and it uses only unlabeled data as pseudo-labels when class probabilities exceed a certain threshold for training. Despite its simplified structure, FixMatch outperformed previous methods [28], [29].

More recently, Zhang et al. proposed FlexMatch [31], which incorporates curriculum pseudo-labeling into FixMatch. This method dynamically adjusts thresholds for different classes to use more informative unlabeled data and their associated pseudo-labels. Zheng et al. [32] proposed SimMatch, which bridges the gaps between the semantic and instance similarities by aligning similarity relationships at both semantic and instance levels for different augmentations. SimMatch uses consistency regularization on both semantic and instance levels, using a labeled memory buffer to fully leverage the data annotations on the instance level.

In this study, inspired by the simple structure of FixMatch, we use both weak and strong augmentations of the spectrograms for consistency regularization using the prediction from the weakly augmented spectrogram as a pseudo-label.

#### 2) SEMI-SUPERVISED REGRESSION

Previous SSR methods use both labeled and unlabeled data to train a model to predict continuous values. Zhou and Li [22] proposed co-training regressors (COREG), in which two models are co-trained to generate pseudo-labels from unlabeled data. However, COREG comes with increased computational costs because of the separate training of the two K-nearest neighbors regression models and reduced efficiency caused by information transfer between the models. Wetzel et al. [23] proposed twin neural network regression (TNNR) that trains the difference between the predictions of two unlabeled data pairs through loop consistency. Although TNNR enables learning from unlabeled data through loop consistency, it requires two or more independent models, increasing computational costs. In contrast to existing methods that rely on two or more models, our approach uses a single model, enhancing intuitive comprehension of the architecture and simplifying the inference steps in the model. Moreover, to enhance the impact of consistency regularization with a single model, we suggest applying augmentations such as weak, strong, and mixup augmentation.

Jean et al. [33] proposed semi-supervised deep kernel learning (SSDKL), which focused on minimizing prediction variance of unlabeled data. SSDKL combines neural networks and kernel methods in deep kernel learning, allowing the model to learn nonlinear relationships within the unlabeled data. Nonetheless, the complexity of deep kernel learning results in high computational costs. Our approach diverges from using kernel methods. Instead, we focus on information maximization of unlabeled data in the embedding space, ensuring that the model learns relationships between embedding vectors, thereby simplifying the overall model structure.

Dai et al. [21] proposed uncertainty consistent variational model ensembling (UCVME), which improves the quality of uncertainty estimates on unlabeled data through a BNN and model ensembles. UCVME allows uncertainty estimation through the BNN; however, it requires multiple samplings for posterior distribution estimation, resulting in high computational costs because of ensembles involving more than one model. Unlike previous studies using ensembles or BNN, our approach mitigates computational costs by leveraging a single model for consistency regularization and information maximization. This study proposes SpecRegMatch, an SSR method that boasts low computational requirements, achieved by using a single model without the need for multiple data samplings.

### B. SPECTROGRAM AUGMENTATION

A spectrogram, a transformed representation of signal data, is widely used in signal processing, audio processing, speech recognition, and various other fields. Spectrogram augmentation, involving various transformations applied to a spectrogram, holds significance in SSL as it enhances the model's robustness by generating diverse instances from unlabeled data [15].

Grollmisch and Cano [27] applied FixMatch to audio classification, improving the performance of semi-supervised classification by incorporating both weak and strong augmentations into spectrograms. They introduced suitable augmentations tailored for spectrograms and utilized SpecAugment [34] for strong augmentation, involving masking both the time and frequency axes along with the information contained in the spectrogram. Inspired by this approach, our proposed augmentation method applies multiplicative noise as a weak augmentation and SpecAugment as a strong augmentation. Additionally, Zhang et al. [35] proposed mixup, an augmentation method generating new training data by linearly combining two samples. Mixup allows the model to exhibit linear behavior between training data, enhancing its robustness in predicting adversarial data. In this study, we apply a mixup augmentation approach by combining weakly and strongly augmented spectrograms. In contrast to Grollmisch and Cano [27], we enhance consistency regularization by minimizing the disparity between predictions generated from the mixup-augmented spectrogram and those obtained from the weakly augmented spectrogram. Furthermore, we introduce information maximization in the embedding space as an additional component.

## III. PROPOSED METHOD

### A. NOTATIONS WITH THE PRELIMINARIES OF INPUT DATA

Let $x_l$ represent the signal data paired with its corresponding target value $y$, which is interior noise levels. In addition, let $x_{ul}$ represent the unlabeled signal data. A spectrogram is obtained by applying a short-time Fourier transform (STFT) to the signal data [36]. The spectrogram represents the signal frequency over time. The STFT is computed by dividing the signal into short overlapping windows and calculating the Fourier transform of each window as follows:

$$STFT\{x(t)\} = \int \left[x(t) * w(t-\tau) * \exp(-j2\pi ft)\right] d\tau, \tag{1}$$

where $x(t)$ represents time-domain signal data, and $w(t-\tau)$ is a window function $.f$, $t$, $j$ are frequency, time, and an imaginary unit, respectively. The labeled signal data $x_l$ are transformed into the labeled spectrogram $s_l$, and the unlabeled data $x_{ul}$ are transformed into the unlabeled spectrogram $s_{ul}$ using the STFT.

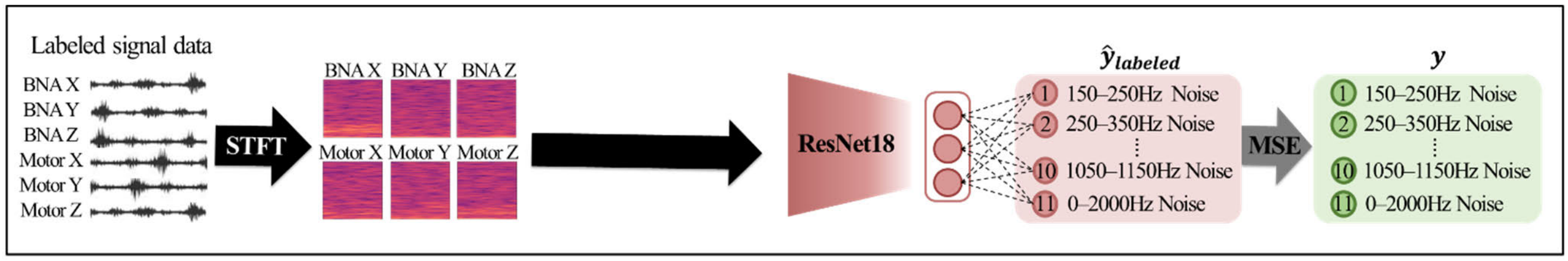

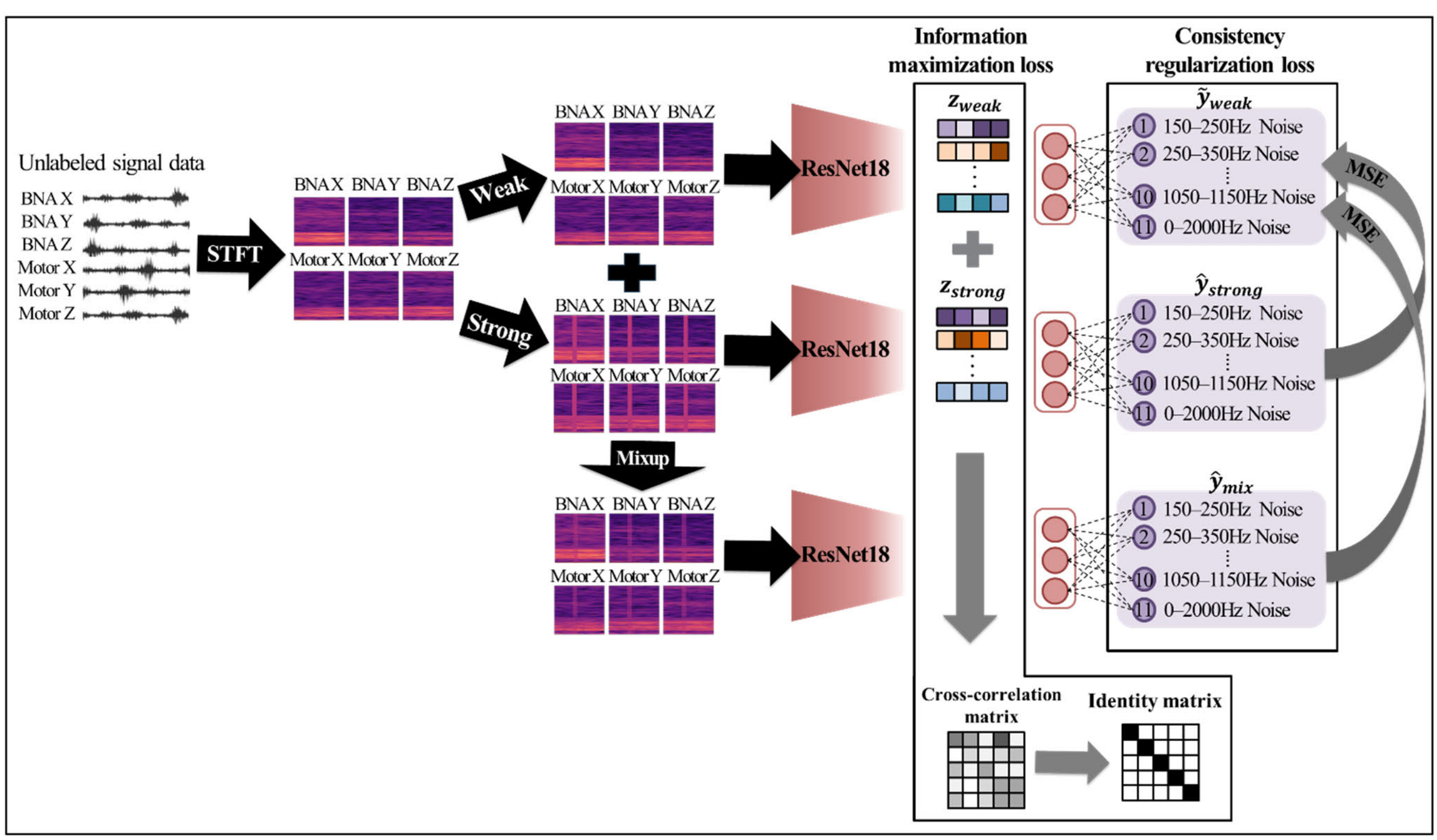

**FIGURE 2.** Architecture of the proposed method, SpecRegMatch.

The proposed approach applies a weak augmentation $A_w$, strong augmentation $A_s$, and mixup augmentation $A_m$ to the unlabeled spectrogram $s_{ul}$. The selection of weak and strong augmentations is informed by Grollmisch and Cano, as discussed in Section II. Weakly augmented spectrograms $s_w$ are generated by applying a weak augmentation $A_w$, which involves random application of multiplicative noise sampled from a uniform distribution between 0.8 and 1.2. For strong augmentation $A_s$, a strongly augmented spectrogram $S_s$ is generated using SpecAugment, which entails masking the time and frequency of the spectrogram. Time masking applies a mask of size $[\mathrm{t_s}, \mathrm{t_s} + \mathrm{t_l}]$ with randomly selected starting points $\mathrm{t_s}$ and a hyperparameter of the time-masking length, denoted as $\mathrm{t_l}$. Similarly, frequency masking applies a mask of size $[f_s, f_s + f_l]$ with a randomly selected frequency starting point $f_s$ and a hyperparameter of the frequency-masking length $f_l$. Both the time-masking length $\mathrm{t_l}$ and frequency-masking length $\mathrm{f_l}$ are set to 20. Mixup augmentation $A_m$ combines the weakly augmented spectrogram $s_w$ and the strongly augmented spectrogram $s_s$ to obtain the following mixup-augmented spectrogram $s_m$:

$$A_m = \alpha (A_w) + (1-\alpha)(A_s) = s_m, \tag{2}$$

where $\alpha$ is a mixup ratio governing the linear combination of the weakly and strongly augmented spectrograms. This ratio $\alpha$ is randomly determined between zero and one through a beta distribution.

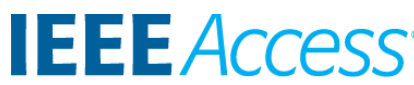

**Algorithm 1** Pseudocode for SpecRegMatch

1: **Input:** a batch of labeled signal data $\mathcal{D}_l = \left\{\left(x_l^i, y^i\right) \mid x_l^i \in X, y^i \in Y\right\}$ and unlabeled signal data $\mathcal{D}_u = \left\{x_{ul}^i \mid x_{ul}^i \in X\right\}$
2: # Apply STFT to the labeled and unlabeled signal data
$S_l = \mathrm{STFT}(D_l), S_{ul} = \mathrm{STFT}(D_u)$
3: # Apply weak, strong, and mixup augmentation to the unlabeled spectrograms
$S_w = A_w(S_{ul}), S_s = A_s(S_{ul}), S_m = A_m(S_w, S_s)$
4: # Forward a labeled spectrogram and unlabeled spectrograms to the backbone network $f(\cdot)$
$Z_{labeled}, Z_{weak}, Z_{strong}, Z_{mix} = f(S_l, S_w, S_s, S_m)$
5: # Forward embedding vectors of the spectrograms to the MLP layer $g(\cdot)$
$\hat{Y}_{labeled}, \tilde{Y}_{weak}, \hat{Y}_{strong}, \hat{Y}_{mix} = g(Z_{labeled}, Z_{weak}, Z_{strong}, Z_{mix})$
6: # Compute supervised loss of labeled data using mean squared error (MSE) function
$\mathcal{L}_{labeled} = MSE\left(\hat{Y}_{labeled}, Y\right)$
7: # Compute consistency regularization loss with two hyperparameters $\lambda_s$ and $\lambda_m$
$\mathcal{L}_{CR} = \lambda_s MSE\left(\hat{Y}_{strong}, \tilde{Y}_{weak}\right) + \lambda_m MSE\left(\hat{Y}_{mix}, \tilde{Y}_{weak}\right)$
8: # Compute information maximization loss with two hyperparameters $\lambda_{IM}$ and $\lambda_r$
$\mathcal{L}_{IM} = \lambda_{IM}\left(\sum_i (1-C_{ii})^2 + \lambda_r \sum_i \sum_{j\neq i} C_{ij}^2\right)$
9: # Compute unsupervised loss of unlabeled data
$\mathcal{L}_{unlabeled} = \mathcal{L}_{CR} + \mathcal{L}_{IM}$
10: # Compute total loss
**return** $\mathcal{L}_{total} = \mathcal{L}_{labeled} + \mathcal{L}_{unlabeled}$

### B. SPECREGMATCH

Figure 2 provides an overview of the proposed SpecRegMatch method, and its pseudocode is presented as Algorithm 1. The model predicts the vehicle interior noise of 11 distinct frequency bands. Given a batch of labeled signal data $\mathcal{D}_l = \left\{\left(x_l^i, y^i\right) \mid x_l^i \in X, y^i \in Y\right\}$ and an equally-sized batch of unlabeled signal data $\mathcal{D}_u = \left\{x_{ul}^i \mid x_{ul}^i \in X\right\}$, where $\mathcal{D}_l$ contains labeled signal data $x_l$ and $\mathcal{D}_u$ contains unlabeled signal data $x_{ul}$. The labeled signal data $\mathcal{D}_l$ and unlabeled signal data $\mathcal{D}_u$ are transformed to labeled spectrograms $S_l = \left\{\left(s_l^i, y^i\right) \mid s_l^i \in S_l, y^i \in Y\right\}$ and unlabeled spectrograms $S_{ul} = \left\{s_{ul}^i \mid s_{ul}^i \in S_{ul}\right\}$ using STFT.

Subsequently, the unlabeled spectrograms $S_{ul}$ are subjected to weak augmentation $A_w$, strong augmentation $A_s$, and mixup augmentation $A_m$, which combines both weakly and strongly augmented spectrograms. Consequently, weakly augmented spectrograms $S_w = \left\{s_w^i \mid s_w^i \in S_w\right\}$, strongly augmented spectrograms $S_s = \left\{s_s^i \mid s_s^i \in S_s\right\}$, and mixup-augmented spectrograms $S_m = \left\{s_m^i \mid s_m^i \in S_m\right\}$ are generated from unlabeled spectrograms $S_{ul}$.

Augmented spectrograms are then transformed into embedding vectors through the backbone network $f(\cdot)$ and subsequently pass through a multilayer perceptron (MLP) $g(\cdot)$ to generate predicted values. The information maximization loss $\mathcal{L}_{IM}$ is computed from these embedding vectors, and the consistency regularization loss $\mathcal{L}_{CR}$ is computed from the predicted values obtained after passing through the MLP.

The overall training loss $\mathcal{L}_{total}$ of SpecRegMatch can be expressed as follows:

$$\mathcal{L}_{total} = \mathcal{L}_{labeled} + \mathcal{L}_{unlabeled}, \tag{3}$$

$$\mathcal{L}_{labeled} = \frac{1}{|S_l|} \sum_{(s_l^i, y^i)\in S_l} \left(y^i - \hat{y}_{labeled}^i\right)^2, \tag{4}$$

$$\mathcal{L}_{unlabeled} = \mathcal{L}_{CR} + \mathcal{L}_{IM}. \tag{5}$$

The supervised loss $\mathcal{L}_{labeled}$ in Equation (4) follows the standard supervised learning conditions. In contrast, the unsupervised loss $\mathcal{L}_{unlabeled}$ combines the consistency regularization loss $\mathcal{L}_{CR}$ and information maximization loss $\mathcal{L}_{IM}$, as shown in Equation (5). For consistency regularization, the model is trained such that prediction of the strongly augmented $\hat{y}_{strong}$ and prediction of the mixup-augmented spectrogram $\hat{y}_{mix}$ closely align with the pseudo-label, represented by prediction of the weakly augmented spectrogram $\tilde{y}_{weak}$. The consistency regularization loss $\mathcal{L}_{CR}$ is expressed as follows:

$$\mathcal{L}_{CR} = \lambda_s \left(\frac{1}{|S_w|} \sum_{(s_w^i)\in S_w} \left(\tilde{y}_{weak}^i - \hat{y}_{strong}^i\right)^2\right) + \lambda_m \left(\frac{1}{|S_w|} \sum_{(s_w^i)\in S_w} \left(\tilde{y}_{weak}^i - \hat{y}_{mix}^i\right)^2\right), \tag{6}$$

$$\tilde{y}_{weak}^i = g\left(f\left(s_w^i\right)\right), \hat{y}_{strong}^i = g\left(f\left(s_s^i\right)\right), \quad \hat{y}_{mix}^i = g\left(f\left(s_m^i\right)\right), \tag{7}$$

where $\lambda_s$ is a hyperparameter that controls the learning influence between prediction of the weakly augmented spectrogram $\tilde{y}_{weak}$ as the pseudo-label and the prediction of the strongly augmented spectrogram $\hat{y}_{strong}$. $\lambda_m$ is a hyperparameter that controls the learning influence between prediction of the weakly augmented spectrogram $\tilde{y}_{weak}$ as the pseudo-label and prediction of the mixup-augmented spectrogram $\hat{y}_{mix}$.

The information maximization loss $\mathcal{L}_{IM}$ comprises invariance loss $\mathcal{L}_{invariance}$ and the reduction loss $\mathcal{L}_{reduction}$, and it is calculated based on the cross-correlation matrix $C$ as follows:

$$\mathcal{L}_{IM} = \lambda_{IM}\left(\mathcal{L}_{invariance} + \lambda_r \mathcal{L}_{reduction}\right) = \lambda_{IM}\left(\sum_i (1-C_{ii})^2 + \lambda_r \sum_i \sum_{j\neq i} C_{ij}^2\right), \tag{8}$$

where $\lambda_{IM}$ is a hyperparameter that controls the overall learning influence of information maximization. Additionally, $\lambda_r$ is a hyperparameter responsible for controlling redundancy. $i$ and $j$ represent the dimension of embedding vectors. Information maximization loss computes the cross-correlation matrix $C$ between augmented samples extracted by embedding vectors from the same data, aiming to make them close to the identity matrix. The cross-correlation matrix $C$ is trained to approximate an identity matrix, with diagonal elements of $C$ set to one to capture invariance between augmented embedding features from the same sample, and off-diagonal elements of $C$ set to zero to reduce the redundancy among different embedding (see Equation (8)). The embedding vectors for both the weakly augmented spectrogram $z_{weak}$ and the strongly augmented spectrogram $z_{strong}$ are extracted from the backbone network $f(\cdot)$. The cross-correlation matrix $C$ is calculated as follows:

$$C_{ij} = \frac{\sum_b z_{weak}^{b,i}, z_{strong}^{b,j}}{\sqrt{\sum_b \left(z_{weak}^{b,i}\right)^2}\sqrt{\sum_b \left(z_{strong}^{b,j}\right)^2}}, \tag{9}$$

where $b$ is the index in the batch samples. $z_{weak}^{b,i}$ denotes the $i$-th embedding feature of the $b$-th data point in the embedding vector of the weakly augmented spectrogram, and $z_{strong}^{b,j}$ represents the $j$-th embedding feature of the $b$-th data in the embedding vector of the strongly augmented spectrogram. The element $C_{ij}$, positioned in the $i$-th row and $j$-th column of the cross-correlation matrix indicates the correlation between two embedding vectors of the weakly and strongly augmented spectrograms $z_{weak}$ and $z_{strong}$. In other words, the cross-correlation matrix quantifies the relationship between the two embedding vectors, providing the information about the similarity of their respective features.

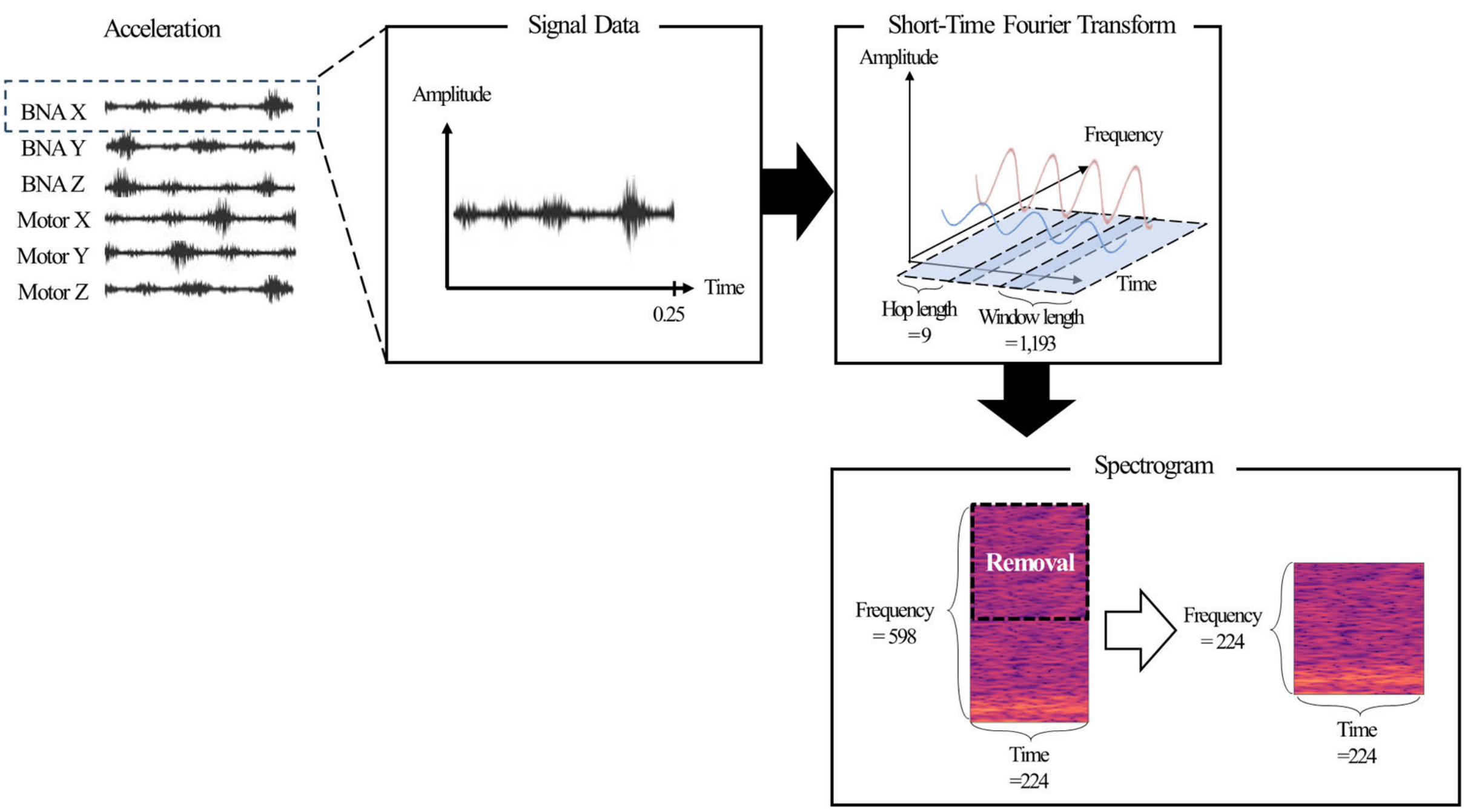

**FIGURE 3.** Pre-processing steps of acceleration data.

## IV. EXPERIMENTS

We used data collected from an electric vehicle equipped with components developed by a Korean auto parts vendor. The dataset was divided into training, validation, and testing sets, as presented in Table 2. The number of data indicates the typical count of data points where the acceleration and interior noise data are matched one-to-one.

To examine the performance with a limited amount of labeled data, experiments were performed for five cases, each associated with different proportions (K%) of labeled data: 2%, 7%, 28%, 52%, and 100%. Dividing the labeled training data into these five proportions was based on expert opinions, which suggested that the training data should be divided according to the collected environmental conditions, such as the collection date.

Figure 3 illustrates the pre-processing steps for the acceleration data. In the first step, the acceleration data were sampled at a rate of 12,800 Hz, and observations were segmented at intervals of 0.25 seconds. The second step involved generating a spectrogram using window and hop lengths of 1,193 and 9, respectively, resulting in a spectrogram size of 224 × 598. The spectrogram was formed by iteratively applying fast Fourier transform to 1,193 accelerometer data points, moving along the time axis with nine data points at a time, and concatenating their values. The 224 on the x-axis of the spectrogram represents the time range, and the 598 on the y-axis represents the frequency range. In the third step, based on expert guidance, frequency bands exceeding 2,400 Hz were deemed unwanted data and subsequently removed. Finally, spectrograms of size 224 × 224, representing the time and frequency for each device (BNA X, BNA Y, motor Z, motor X, motor Y, and motor Z), were used in the experiment. The target variable, representing vehicle interior noise in 11 frequency bands, was directly measured inside the vehicle at 0.25-second intervals within the following 11 frequency ranges: 150–250 Hz, 250–350 Hz, 350–450 Hz, 450–550 Hz, 550–650 Hz, 650–750 Hz, 750–850 Hz, 850–950 Hz, 950–1,050 Hz, 1,050–1,150 Hz, and 0–2,000 Hz. Based on expert opinions, data exceeding the threshold in the 0–2,000 Hz frequency band were considered outliers and consequently excluded, along with the spectrogram of acceleration data.

**TABLE 2. Training, validation, and testing data in the experiment dataset. K represents the proportion of labeled data.**

| Training data | | Validation data | Testing data |
|---|---|---|---|
| Labeled data | Unlabeled data | | |
| Number of data (K %) | Number of data | Number of data | Number of data |
| 242 (2%) | 15,479 | 3,266 | 3,490 |
| 870 (7%) | | | |
| 3,396 (28%) | | | |
| 6,354 (52%) | | | |
| 12,257 (100%) | | | |

**TABLE 3. Details of the hyperparameter search. The selected optimal hyperparameters are shown in bold.**

| Shared | Searching Value |
|---|---|
| Optimizer | AdamW |
| Training epochs | 50 |
| Warm-up training epochs in linear ramp-up function | {10, 15, **20**, 30} |
| Learning rate | {0.0001, 0.0005, **0.001**} |
| Batch size | 128 |
| **SSDKL** | |
| Coefficient of variance loss | {1, **5**, 10} |
| **UCVME** | |
| Number of samplings for variational inference | 5 |
| Dropout probability of backbone network | {1%, **5%**, 10%} |
| Weighting parameter for unlabeled data | {**5**, 10} |
| **SpecRegMatch (proposed)** | |
| Coefficient of weak and strong augmented spectrogram predictions ($\lambda_s$) | {0.001, 0.005, 0.01, **0.05**, 0.1} |
| Coefficient of weak and mixup augmented spectrogram predictions ($\lambda_m$) | {0.001, 0.005, 0.01, **0.1**, 0.5} |
| Coefficient of information maximization loss ($\lambda_{IM}$) | {0.01, 0.001, **0.005**, 0.001} |
| Coefficient of reduction loss ($\lambda_r$) | {**0.0001**, 0.005, 0.001} |

To demonstrate the superiority of the proposed SpecRegMatch, we conducted a performance comparison with the four SSR methods described in Section II. We evaluated the proposed method based on the individual scores, mean scores of individual scores, and standard deviations for the 11 targets. Mean absolute error (MAE) and $R^2$ were used for assessment. Additionally, we measured floating point operations (FLOPs) to demonstrate that SpecRegMatch has lower computational costs compared to other SSR methods.

FLOPs were measured based on one epoch, with a notation unit of $10^{13}$. For a backbone model, we used the ResNet18 widely used for spectrograms [3], [4], [5], [6], [7], [8], [9], [10], [11], [12], [13], [14], [15], [16], [17], [18], [19], [20], [21], [22], [23], [24], [25], [26], [27], [28], [29], [30], [31], [32], [33], [34], [35], [36], [37], [38], [39]. The MLP layer comprises two linear layers with batch normalization in between. Each layer has 512 and 11 output dimensions, respectively. For each method, Table 3 outlines the details of the hyperparameters investigated in this study. The optimal hyperparameters were chosen to attain the highest validation score.

Table 4 presents the average and standard deviation of $R^2$ obtained from seven repeated experiments for the labeled data at five different rates. We conducted experiments to ensure the consistency of each method in both the MAE and $R^2$ metrics. However, to streamline the contents of the table, we only show $R^2$ in Table 4. A table containing MAE can be found in Table 5 of Appendix. In a scenario with five different labeling rates, we compared the performance of six methods for 11 targets. The best performance among the six methods is highlighted in bold, and the second-best performance was underlined. When analyzing the mean scores of the 11 targets, all SSR methods performed better than the baseline (supervised learning) at label rates (K) of 2%, 7%, and 28%. Among the SSR methods, the proposed SpecRegMatch performed the best across all labeled data rates compared with the other four SSR methods. Specifically, at a labeling rate (K) of 2% with the least labeled data, the mean $R^2$ of the baseline was $-0.555$, while the $R^2$ of the proposed method was 0.484, demonstrating the superiority of the proposed method even with a limited amount of labeled data.

Furthermore, SpecRegMatch demonstrated high performance compared to the second-best performing TNNR, achieving an $R^2$ of 0.458. When evaluating the individual target $R^2$ in addition to the mean $R^2$, the proposed method achieved the highest number of individual best performances compared to the other methods, five out of 11 for at a 2% labeling rate, six at 7%, nine at 28%, ten at 52%, and nine at 100%. In addition, it is worth noting that the proposed method achieved superior results using only 28% of the training labels (K=28%) when compared to fully supervised (K=100%) results. These experiment results clearly demonstrate the effectiveness of SpecRegMatch in reducing dependence on labeled data, while consistently achieving high performance.

Given the sizes of the labeled and unlabeled data as $m$ and $n$, respectively, the time complexity of model training, expressed in big-O notation, for COREG, SSDKL, TNNR, and SpecRegMatch can be denoted as $O\left((m+n)^2\right)$. However, for UCVME, because of the multiple sampling of unlabeled data at each iteration, the time complexity can be represented as $O\left((m+n)\left(m+n^2\right)\right)$. Figure 4 shows that SpecRegMatch demonstrated the lowest value, with $10^{13}$

**TABLE 4. Average and standard deviation (in parentheses) of $R^2$ results under five different rates (K%) of labeled data. The best results are in bold, and the second-best results are underlined for each labeled data rate. The Wilcoxon rank-sum test is used to verify significant differences between the best and second-best mean results, and is noted by p-value (*: $p < 0.05$).**

| The rate (K%) of labeled data | Method | 150–250 Hz | 250–350 Hz | 350–450 Hz | 450–550 Hz | 550–650 Hz | 650–750 Hz | 750–850 Hz | 850–950 Hz | 950–1,050 Hz | 1,050–1,150 Hz | 0–2,000 Hz | Mean |
|---|---|---|---|---|---|---|---|---|---|---|---|---|---|
| 2% | Baseline (supervised) | -1.418 (1.097) | -0.553 (0.583) | -0.258 (0.779) | -0.754 (0.976) | -0.711 (1.666) | -0.551 (1.110) | -0.407 (1.040) | -0.070 (0.423) | -0.074 (0.477) | -0.397 (0.636) | -0.910 (1.070) | -0.555 (0.547) |
| | COREG | 0.333 (0.281) | 0.273 (0.226) | **0.824 (0.027)** | 0.690 (0.066) | 0.590 (0.051) | 0.449 (0.120) | 0.540 (0.193) | **0.463 (0.141)** | 0.082 (0.258) | 0.058 (0.162) | 0.597 (0.201) | 0.445 (0.060) |
| | SSDKL | **0.528 (0.092)** | 0.322 (0.120) | 0.704 (0.068) | 0.692 (0.02) | 0.484 (0.096) | 0.377 (0.085) | 0.488 (0.045) | 0.399 (0.048) | 0.179 (0.061) | 0.010 (0.056) | 0.522 (0.070) | 0.428 (0.032) |
| | TNNR | 0.514 (0.140) | 0.302 (0.341) | 0.739 (0.091) | 0.724 (0.067) | 0.573 (0.107) | 0.386 (0.180) | **0.640 (0.063)** | 0.368 (0.153) | 0.050 (0.313) | 0.074 (0.121) | **0.672 (0.058)** | 0.458 (0.026) |
| | UCVME | 0.320 (0.180) | 0.069 (0.161) | 0.783 (0.037) | 0.677 (0.052) | 0.582 (0.085) | **0.502 (0.081)** | 0.586 (0.061) | 0.370 (0.169) | **0.208 (0.086)** | 0.003 (0.159) | 0.443 (0.145) | 0.413 (0.079) |
| | SpecRegMatch (proposed) | 0.481 (0.110) | **0.390 (0.127)** | 0.768 (0.050) | **0.747 (0.040)** | **0.630 (0.066)** | **0.502 (0.091)** | 0.577 (0.078) | 0.379 (0.116) | 0.190 (0.121) | **0.076 (0.104)** | 0.580 (0.060) | **0.484* (0.034)** |
| 7% | Baseline (supervised) | 0.561 (0.085) | 0.433 (0.140) | 0.832 (0.032) | 0.63 (0.194) | 0.706 (0.032) | 0.712 (0.058) | 0.718 (0.104) | 0.675 (0.042) | 0.588 (0.061) | 0.480 (0.176) | 0.612 (0.081) | 0.631 (0.033) |
| | COREG | 0.683 (0.074) | 0.464 (0.105) | 0.809 (0.089) | 0.779 (0.046) | 0.702 (0.069) | 0.728 (0.057) | 0.737 (0.091) | 0.679 (0.066) | **0.610 (0.046)** | 0.531 (0.056) | 0.723 (0.064) | 0.677 (0.051) |
| | SSDKL | **0.697 (0.022)** | 0.491 (0.053) | 0.839 (0.016) | 0.729 (0.05) | 0.739 (0.019) | 0.707 (0.031) | 0.717 (0.023) | 0.672 (0.032) | 0.576 (0.034) | 0.512 (0.043) | 0.731 (0.028) | 0.674 (0.015) |
| | TNNR | 0.587 (0.136) | 0.482 (0.183) | 0.843 (0.047) | **0.785 (0.033)** | 0.686 (0.057) | 0.682 (0.048) | 0.729 (0.026) | 0.666 (0.03) | 0.560 (0.032) | 0.380 (0.122) | 0.675 (0.099) | 0.643 (0.036) |
| | UCVME | 0.537 (0.323) | 0.402 (0.363) | 0.840 (0.051) | 0.615 (0.253) | **0.769 (0.022)** | 0.715 (0.100) | 0.743 (0.059) | **0.691 (0.054)** | 0.578 (0.096) | 0.562 (0.042) | 0.593 (0.334) | 0.640 (0.150) |
| | SpecRegMatch (proposed) | 0.669 (0.029) | **0.550 (0.041)** | **0.845 (0.022)** | 0.775 (0.031) | 0.707 (0.028) | **0.741 (0.019)** | **0.751 (0.024)** | 0.680 (0.030) | 0.583 (0.031) | **0.581 (0.031)** | **0.750 (0.016)** | **0.694 (0.005)** |
| 28% | Baseline (supervised) | 0.735 (0.048) | 0.705 (0.042) | 0.841 (0.033) | 0.790 (0.053) | 0.687 (0.091) | 0.755 (0.031) | 0.753 (0.03) | 0.677 (0.052) | 0.553 (0.136) | 0.539 (0.045) | 0.798 (0.032) | 0.712 (0.025) |
| | COREG | 0.769 (0.029) | **0.745 (0.026)** | 0.871 (0.024) | **0.845 (0.031)** | 0.775 (0.026) | 0.768 (0.023) | 0.778 (0.024) | 0.728 (0.034) | 0.627 (0.031) | 0.576 (0.046) | 0.813 (0.020) | 0.754 (0.008) |
| | SSDKL | 0.740 (0.08) | 0.714 (0.052) | 0.830 (0.12) | 0.813 (0.081) | 0.772 (0.016) | 0.764 (0.032) | 0.766 (0.044) | 0.677 (0.069) | 0.628 (0.055) | 0.535 (0.052) | 0.790 (0.080) | 0.730 (0.059) |
| | TNNR | 0.725 (0.087) | 0.729 (0.031) | 0.833 (0.039) | 0.828 (0.027) | 0.763 (0.031) | 0.769 (0.019) | 0.780 (0.017) | 0.712 (0.011) | 0.644 (0.027) | 0.523 (0.036) | 0.780 (0.072) | 0.735 (0.019) |
| | UCVME | 0.760 (0.020) | 0.738 (0.032) | 0.874 (0.013) | 0.822 (0.015) | **0.805 (0.009)** | 0.791 (0.011) | 0.802 (0.012) | 0.738 (0.021) | 0.637 (0.019) | 0.568 (0.031) | 0.818 (0.022) | 0.759 (0.010) |
| | SpecRegMatch (proposed) | **0.791 (0.016)** | **0.745 (0.036)** | **0.878 (0.010)** | 0.833 (0.026) | 0.791 (0.021) | **0.799 (0.010)** | **0.807 (0.015)** | **0.755 (0.014)** | **0.670 (0.021)** | **0.579 (0.014)** | **0.826 (0.015)** | **0.770 (0.007)** |
| 52% | Baseline (supervised) | 0.784 (0.044) | 0.646 (0.140) | 0.875 (0.020) | 0.840 (0.030) | 0.747 (0.036) | 0.753 (0.087) | 0.796 (0.026) | 0.675 (0.091) | 0.669 (0.024) | 0.549 (0.115) | **0.846 (0.014)** | 0.744 (0.017) |
| | COREG | 0.734 (0.044) | 0.749 (0.043) | 0.881 (0.012) | 0.856 (0.027) | 0.784 (0.031) | 0.778 (0.022) | 0.778 (0.036) | 0.729 (0.029) | 0.644 (0.021) | 0.579 (0.067) | 0.774 (0.067) | 0.753 (0.012) |
| | SSDKL | 0.717 (0.116) | 0.696 (0.059) | 0.812 (0.111) | 0.816 (0.062) | 0.753 (0.036) | 0.769 (0.059) | 0.762 (0.069) | 0.698 (0.075) | 0.640 (0.076) | 0.605 (0.057) | 0.757 (0.122) | 0.730 (0.074) |
| | TNNR | 0.688 (0.128) | 0.711 (0.062) | 0.841 (0.028) | 0.818 (0.042) | 0.759 (0.022) | 0.773 (0.021) | 0.770 (0.058) | 0.702 (0.028) | 0.65 (0.054) | 0.558 (0.081) | 0.806 (0.035) | 0.734 (0.017) |
| | UCVME | 0.746 (0.058) | 0.754 (0.021) | 0.870 (0.025) | 0.841 (0.017) | 0.789 (0.017) | 0.790 (0.017) | 0.796 (0.021) | 0.736 (0.029) | 0.673 (0.02) | 0.608 (0.053) | 0.807 (0.033) | 0.765 (0.023) |
| | SpecRegMatch (proposed) | **0.800 (0.013)** | **0.769 (0.030)** | **0.887 (0.005)** | **0.866 (0.009)** | **0.793 (0.015)** | **0.806 (0.009)** | **0.811 (0.020)** | **0.761 (0.009)** | **0.707 (0.019)** | **0.619 (0.020)** | 0.840 (0.013) | **0.787* (0.008)** |
| 100% | Baseline (supervised) | 0.809 (0.030) | 0.748 (0.057) | 0.883 (0.010) | 0.852 (0.034) | 0.788 (0.030) | 0.774 (0.040) | 0.778 (0.021) | 0.638 (0.097) | 0.593 (0.04) | 0.497 (0.146) | 0.848 (0.025) | 0.746 (0.024) |
| | COREG | 0.758 (0.037) | 0.758 (0.049) | 0.889 (0.010) | 0.858 (0.017) | **0.813 (0.012)** | 0.806 (0.014) | 0.803 (0.020) | 0.717 (0.017) | **0.679 (0.03)** | **0.630 (0.026)** | 0.843 (0.021) | 0.778 (0.013) |
| | SSDKL | 0.791 (0.02) | 0.738 (0.02) | 0.862 (0.01) | 0.847 (0.017) | 0.772 (0.014) | 0.782 (0.010) | 0.785 (0.011) | 0.72 (0.006) | 0.639 (0.017) | 0.592 (0.026) | 0.834 (0.017) | 0.760 (0.009) |
| | TNNR | 0.703 (0.036) | 0.708 (0.062) | 0.838 (0.042) | 0.820 (0.027) | 0.763 (0.029) | 0.729 (0.082) | 0.710 (0.074) | 0.651 (0.077) | 0.624 (0.052) | 0.573 (0.06) | 0.742 (0.159) | 0.715 (0.035) |
| | UCVME | 0.713 (0.104) | 0.712 (0.127) | 0.840 (0.072) | 0.829 (0.048) | 0.734 (0.101) | 0.767 (0.083) | 0.771 (0.042) | 0.724 (0.030) | 0.645 (0.018) | 0.610 (0.039) | 0.814 (0.035) | 0.742 (0.052) |
| | SpecRegMatch (proposed) | **0.840 (0.012)** | **0.772 (0.029)** | **0.891 (0.014)** | **0.880 (0.010)** | **0.813 (0.010)** | **0.813 (0.009)** | **0.819 (0.008)** | **0.733 (0.016)** | 0.639 (0.025) | 0.629 (0.024) | **0.862 (0.012)** | **0.790* (0.005)** |

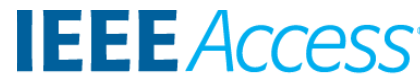

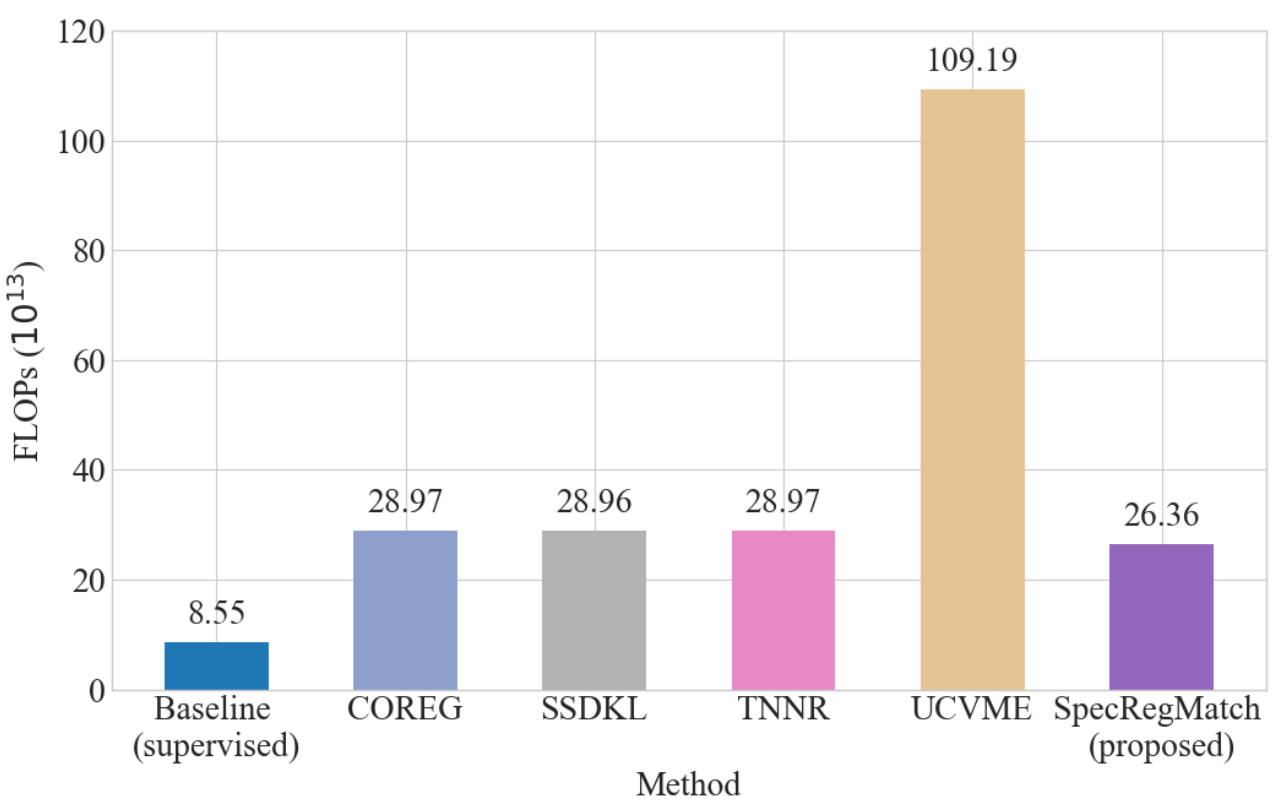

**FIGURE 4.** Comparison of computational cost, measured in $10^{13}$ FLOPs, among various SSR methods and supervised learning method.

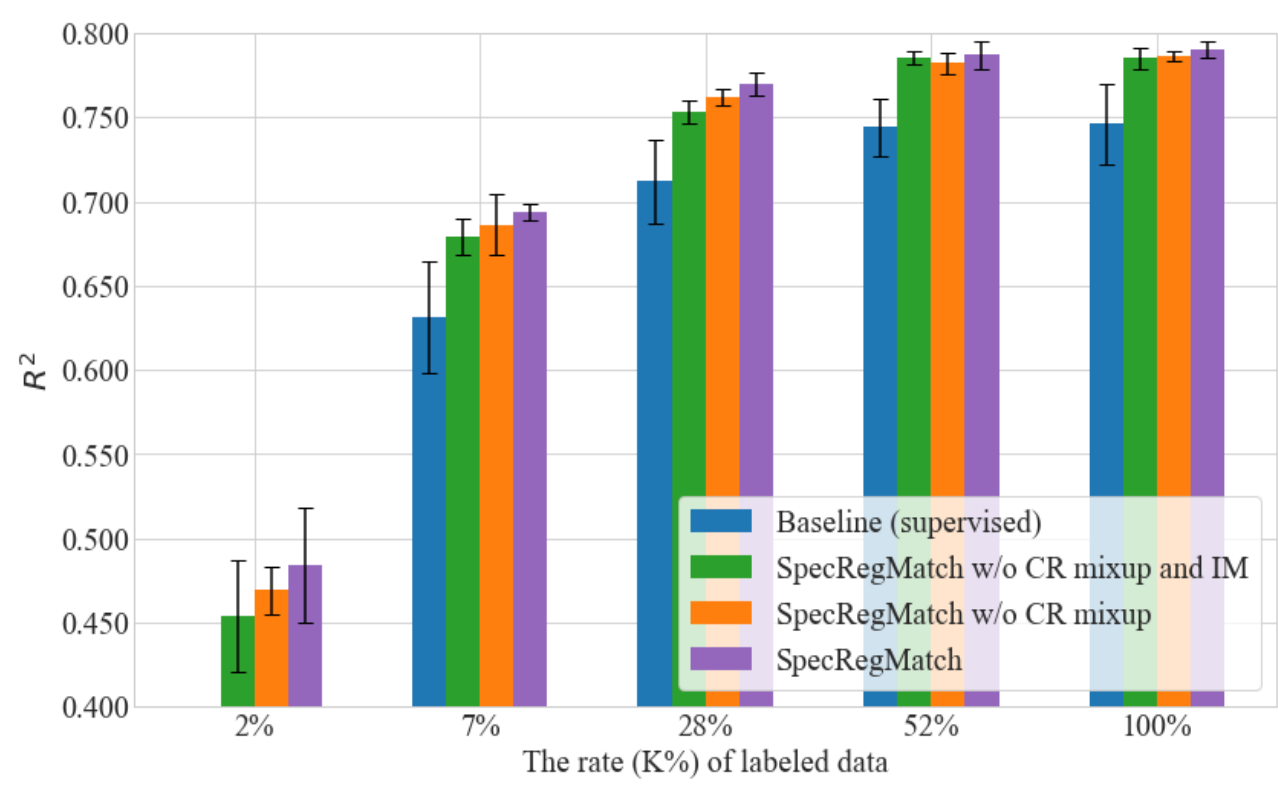

**FIGURE 5.** Results of ablation study for mean $R^2$ of 11 targets under five different rates (K%) of labeled data. CR and IM represent consistency regularization loss and information maximization loss, respectively. When rate K is 2%, $R^2$ for the baseline is negative and therefore not shown.

FLOPs totaling 26.36 among all SSR methods, even while achieving high performance. This computational efficiency is particularly striking when contrasted with UCVME, which, because of multiple samplings and ensemble models, exhibited the highest FLOPs at $10^{13}$ FLOPs of 109.19. While big-O notation provides an upper bound on time complexity, which is the same for SSR methods except UCVME, the proposed approach demonstrated competitiveness when examining the FLOPs results on real computers. These results demonstrate the computational efficiency of SpecRegMatch, which uses a single model to achieve high performance with the lowest FLOPs compared to other SSR methods.

Figure 5 illustrates ablation studies that examine the contribution of each component to the proposed method. Four modules were compared: the baseline (supervised), SpecRegMatch without mixup augmentation of consistency regularization and information maximization losses (SpecRegMatch w/o CR mixup and IM), SpecRegMatch without mixup augmentation of consistency regularization loss (SpecRegMatch w/o CR mixup), and proposed SpecRegMatch. Experiments were performed for five different rates (K%) of labeled data, and the mean $R^2$ values from seven repeated experiments of the 11 targets were used for evaluation. At the most limited labeled data rate (K=2%), the significant difference between the baseline and SpecRegMatch without CR mixup and IM demonstrated the importance of the consistency regularization loss using both strong and strong augmentations.

Furthermore, the difference between SpecRegMatch w/o CR mixup and IM and SpecRegMatch w/o CR mixup indicated that the information maximization loss influenced the performance. The difference between SpecRegMatch w/o CR mixup and SpecRegMatch demonstrated that using mixup augmentation for consistency regularization loss improved the performance. Each component contributed significantly at relatively limited labeled data rates (2%, 7%, and 28%). Conversely, at higher labeled rates (52% and 100%), the component related to consistency regularization, using both weakly and strongly augmentations, continued to contribute significantly; however, other components exhibited lower contributions compared to those at the limited labeled data rates (2%, 7%, and 28%). These results demonstrated the benefit of each component, contributing significantly to model training in situations with limited labeled data. The results also demonstrate that consistency regularization can be enhanced by adding mixup augmentation. Finally, the results indicated that combining consistency regularization with information maximization by using various augmentations to the embedding vectors and predicted values enables robust training of the model even with a single model.

## V. CONCLUSION

This study introduces SpecRegMatch, an SSR method designed to predict vehicle interior noise levels based on a spectrogram. SpecRegMatch capitalizes on consistency regularization and information maximization within a unified model, thereby reducing computational costs in comparison to preceding SSR methods. Experimental findings showcase that SpecRegMatch achieves state-of-the-art results when compared to other SSR methods, particularly in scenarios with limited labeled data. Ablation studies underscore the significance of SpecRegMatch's components, including consistency regularization, information maximization, and mixup augmentation, all of which contribute to the improved performance of the model. This advancement in SSR holds promise for adaptation across diverse domains reliant on signal data.

In practical industry applications, aiming to replace existing programs that manage vehicle interior noise using deep learning, a more precise prediction of noise at finer frequency units is desirable. While the frequency intervals measured in this study are at 100Hz intervals, it is noteworthy that the program used by a Korean automotive vendor for measuring and managing vehicle noise operates at 10Hz intervals. Despite this disparity, our approach is significant because it addresses the need for more detailed frequency predictions, in contrast to existing research that predominantly focuses on predicting the overall frequency of vehicle interior noise.

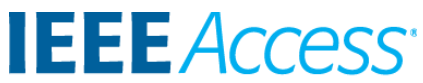

**TABLE 5.** Average and standard deviations (in parentheses) of MAE results under five different rates (K%) of labeled data. The best results are in bold, and the second-best results are underlined for each labeled data rate. The Wilcoxon rank-sum test is used to verify significant differences between the best and second-best mean results, and is noted by p-value (*: $p < 0.05$).

| The rate (K%) of labeled data | Method | 150–250 Hz | 250–350 Hz | 350–450 Hz | 450–550 Hz | 550–650 Hz | 650–750 Hz | 750–850 Hz | 850–950 Hz | 950–1,050 Hz | 1,050–1,150 Hz | 0–2,000 Hz | Mean |
|---|---|---|---|---|---|---|---|---|---|---|---|---|---|
| 2% | Baseline (supervised) | 5.469 (1.763) | 3.745 (0.664) | 5.472 (2.143) | 6.687 (2.126) | 5.260 (1.542) | 5.167 (1.370) | 5.489 (1.449) | 4.244 (0.791) | 4.402 (0.851) | 4.227 (0.702) | 3.825 (1.381) | 4.908 (0.968) |
| | COREG | 2.814 (0.688) | 2.926 (0.530) | **1.992 (0.139)** | 2.640 (0.318) | 2.910 (0.092) | 3.305 (0.217) | 3.047 (0.397) | **2.958 (0.301)** | 4.046 (0.482) | 3.623 (0.239) | 1.989 (0.575) | 2.932 (0.152) |
| | SSDKL | 2.485 (0.280) | 2.844 (0.317) | 2.563 (0.325) | 2.629 (0.081) | 3.204 (0.287) | 3.469 (0.154) | 3.404 (0.114) | 3.230 (0.141) | 4.001 (0.139) | 3.699 (0.165) | 2.162 (0.241) | 3.063 (0.070) |
| | TNNR | **2.415 (0.242)** | 2.768 (0.683) | 2.425 (0.413) | 2.510 (0.287) | 2.960 (0.331) | 3.547 (0.480) | **2.887 (0.166)** | 3.280 (0.313) | 4.188 (0.564) | 3.719 (0.318) | **1.817 (0.225)** | 2.956 (0.065) |
| | UCVME | 3.113 (0.516) | 3.417 (0.337) | 2.175 (0.140) | 2.691 (0.239) | 2.996 (0.369) | **3.182 (0.231)** | 3.130 (0.264) | 3.184 (0.238) | 3.937 (0.195) | 3.744 (0.255) | 2.530 (0.418) | 3.100 (0.238) |
| | SpecRegMatch (proposed) | 2.622 (0.319) | **2.661 (0.313)** | 2.264 (0.231) | **2.398 (0.188)** | **2.772 (0.282)** | 3.223 (0.282) | 3.140 (0.254) | 3.348 (0.256) | **3.871 (0.197)** | **3.550 (0.230)** | 2.067 (0.244) | **2.901 (0.092)** |
| 7% | Baseline (supervised) | 2.436 (0.237) | 2.542 (0.342) | 1.937 (0.196) | 2.942 (0.787) | 2.446 (0.175) | 2.448 (0.177) | 2.459 (0.281) | 2.395 (0.150) | 2.757 (0.230) | 2.561 (0.520) | 2.046 (0.233) | 2.452 (0.106) |
| | COREG | 2.025 (0.26) | 2.353 (0.188) | 2.065 (0.523) | 2.245 (0.234) | 2.475 (0.326) | 2.406 (0.276) | 2.456 (0.500) | 2.366 (0.298) | **2.633 (0.188)** | 2.369 (0.155) | 1.617 (0.261) | 2.274 (0.248) |
| | SSDKL | **1.967 (0.102)** | 2.417 (0.165) | 1.848 (0.055) | 2.468 (0.249) | 2.283 (0.076) | 2.411 (0.106) | 2.478 (0.058) | **2.341 (0.077)** | 2.780 (0.116) | 2.447 (0.141) | 1.577 (0.127) | 2.274 (0.037) |
| | TNNR | 2.200 (0.284) | 2.256 (0.316) | **1.842 (0.220)** | **2.171 (0.158)** | 2.565 (0.255) | 2.576 (0.200) | 2.482 (0.18) | 2.428 (0.136) | 2.832 (0.113) | 2.742 (0.272) | 1.665 (0.259) | 2.342 (0.104) |
| | UCVME | 2.421 (0.966) | 2.573 (0.853) | 1.886 (0.342) | 2.957 (1.035) | **2.152 (0.118)** | 2.489 (0.540) | 2.469 (0.373) | 2.349 (0.249) | 2.766 (0.397) | 2.299 (0.139) | 1.947 (0.973) | 2.392 (0.533) |
| | SpecRegMatch (proposed) | 2.050 (0.114) | **2.211 (0.107)** | 1.872 (0.142) | 2.298 (0.186) | 2.483 (0.124) | **2.398 (0.108)** | **2.427 (0.132)** | 2.422 (0.114) | 2.722 (0.093) | **2.257 (0.110)** | **1.523 (0.070)** | **2.242 (0.023)** |
| 28% | Baseline (supervised) | 1.844 (0.183) | 1.760 (0.133) | 1.898 (0.206) | 2.176 (0.306) | 2.517 (0.35) | 2.281 (0.134) | 2.356 (0.155) | 2.397 (0.217) | 2.812 (0.291) | 2.384 (0.141) | 1.345 (0.160) | 2.161 (0.108) |
| | COREG | 1.696 (0.087) | 1.641 (0.063) | 1.710 (0.173) | 1.861 (0.188) | 2.146 (0.107) | 2.211 (0.161) | 2.184 (0.168) | 2.174 (0.160) | 2.621 (0.107) | 2.261 (0.120) | 1.245 (0.094) | 1.977 (0.055) |
| | SSDKL | 1.690 (0.085) | 1.691 (0.069) | 1.669 (0.106) | **1.850 (0.09)** | 2.125 (0.083) | 2.128 (0.047) | 2.170 (0.108) | 2.288 (0.116) | 2.525 (0.111) | 2.279 (0.03) | **1.205 (0.060)** | 1.965 (0.050) |
| | TNNR | 1.824 (0.203) | 1.723 (0.085) | 1.906 (0.236) | 1.954 (0.168) | 2.224 (0.173) | 2.211 (0.121) | 2.220 (0.109) | 2.257 (0.071) | 2.561 (0.105) | 2.488 (0.164) | 1.336 (0.182) | 2.064 (0.068) |
| | UCVME | 1.702 (0.062) | 1.655 (0.066) | 1.672 (0.062) | 1.985 (0.102) | **1.984 (0.062)** | 2.064 (0.059) | 2.045 (0.059) | 2.082 (0.079) | 2.550 (0.066) | 2.254 (0.081) | 1.212 (0.084) | 1.928 (0.024) |
| | SpecRegMatch (proposed) | **1.642 (0.063)** | **1.631 (0.110)** | **1.654 (0.071)** | 1.938 (0.169) | 2.102 (0.125) | **2.051 (0.075)** | **2.037 (0.090)** | **2.026 (0.076)** | **2.443 (0.098)** | **2.244 (0.060)** | 1.255 (0.104) | **1.911 (0.047)** |
| 52% | Baseline (supervised) | 1.670 (0.175) | 1.911 (0.431) | 1.664 (0.108) | 1.875 (0.175) | 2.262 (0.143) | 2.213 (0.218) | 2.148 (0.181) | 2.349 (0.444) | 2.461 (0.113) | 2.369 (0.413) | **1.162 (0.087)** | 2.008 (0.092) |
| | COREG | 1.887 (0.171) | 1.629 (0.139) | 1.637 (0.094) | 1.765 (0.130) | 2.050 (0.128) | 2.158 (0.102) | 2.223 (0.231) | 2.202 (0.152) | 2.529 (0.102) | 2.223 (0.173) | 1.357 (0.199) | 1.969 (0.066) |
| | SSDKL | 1.803 (0.183) | 1.793 (0.131) | 1.842 (0.155) | 1.910 (0.191) | 2.219 (0.115) | 2.127 (0.116) | 2.228 (0.129) | 2.274 (0.185) | 2.532 (0.233) | 2.152 (0.141) | 1.327 (0.159) | 2.019 (0.122) |
| | TNNR | 2.037 (0.471) | 1.718 (0.133) | 1.898 (0.152) | 2.039 (0.254) | 2.223 (0.114) | 2.235 (0.128) | 2.222 (0.194) | 2.272 (0.149) | 2.527 (0.216) | 2.304 (0.183) | 1.279 (0.115) | 2.069 (0.058) |
| | UCVME | 1.819 (0.256) | 1.621 (0.078) | 1.692 (0.157) | 1.859 (0.089) | **2.048 (0.084)** | 2.065 (0.074) | 2.088 (0.11) | 2.128 (0.138) | 2.395 (0.076) | **2.098 (0.105)** | 1.238 (0.125) | 1.914 (0.090) |
| | SpecRegMatch (proposed) | **1.614 (0.043)** | **1.574 (0.106)** | **1.600 (0.033)** | **1.730 (0.062)** | 2.081 (0.075) | **2.032 (0.067)** | **2.051 (0.153)** | **2.031 (0.033)** | **2.297 (0.091)** | 2.193 (0.085) | 1.173 (0.073) | **1.852* (0.042)** |
| 100% | Baseline (supervised) | 1.557 (0.122) | 1.662 (0.208) | 1.634 (0.078) | 1.829 (0.212) | 2.111 (0.185) | 2.205 (0.207) | 2.216 (0.092) | 2.436 (0.292) | 2.739 (0.157) | 2.514 (0.358) | 1.113 (0.092) | 2.001 (0.083) |
| | COREG | 1.842 (0.171) | 1.615 (0.185) | 1.575 (0.077) | 1.788 (0.122) | **1.942 (0.073)** | **1.979 (0.111)** | 2.076 (0.126) | 2.116 (0.058) | **2.382 (0.126)** | **2.079 (0.104)** | 1.163 (0.140) | 1.869 (0.064) |
| | SSDKL | 1.633 (0.088) | 1.688 (0.074) | 1.729 (0.063) | 1.826 (0.100) | 2.163 (0.080) | 2.126 (0.061) | 2.175 (0.095) | 2.201 (0.021) | 2.602 (0.102) | 2.228 (0.123) | 1.130 (0.059) | 1.955 (0.046) |
| | TNNR | 1.954 (0.197) | 1.735 (0.131) | 1.922 (0.259) | 1.987 (0.151) | 2.175 (0.146) | 2.370 (0.310) | 2.536 (0.259) | 2.442 (0.374) | 2.633 (0.232) | 2.300 (0.201) | 1.367 (0.297) | 2.129 (0.124) |
| | UCVME | 1.948 (0.354) | 1.736 (0.399) | 1.865 (0.423) | 1.947 (0.281) | 2.312 (0.448) | 2.200 (0.392) | 2.270 (0.234) | 2.155 (0.115) | 2.509 (0.094) | 2.171 (0.133) | 1.279 (0.186) | 2.036 (0.236) |
| | SpecRegMatch (proposed) | **1.434 (0.056)** | **1.557 (0.101)** | **1.547 (0.089)** | **1.639 (0.073)** | 1.970 (0.068) | 1.985 (0.060) | **1.998 (0.057)** | **2.107 (0.063)** | 2.569 (0.122) | 2.159 (0.099) | **1.027 (0.051)** | **1.817* (0.022)** |

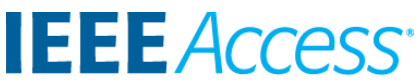

In future work, we aim to explore approaches to improve performance when training a model with a multi-output structure, especially as we increase the number of outputs (frequency bands of vehicle interior noise). Specifically, we plan to investigate techniques that strategically weight output variables within an existing multi-output structure, taking into account the inherent relationships between output variables. In addition, we will explore the use of a threshold to select reliable predictions for unlabeled samples to improve the robustness and stability of the SSR method. This is because shifts in data distributions between labeled and unlabeled data in real-world scenarios may lead to unreliable predictions for unlabeled samples.

## APPENDIX

## EXPERIMENT RESULTS OF MAE

Table 5 presents the MAE results, following the same format as Table 4. Comparing the MAE mean scores of the 11 targets on the labeled data at five different rates, it is again clear that SpecRegMatch outperforms the other SSR methods. Furthermore, when evaluating the MAE on each frequency band, the proposed method demonstrated the highest number of top performances compared to the other methods. For example, it achieved the best results for five out of 11 at a 2% labeling rate (K), five at 7%, eight at 28%, eight at 52%, and seven at 100%.

## REFERENCES

[1] O. B. Sezer, M. U. Gudelek, and A. M. Ozbayoglu, "Financial time series forecasting with deep learning: A systematic literature review: 2005–2019," *Appl. Soft Comput.*, vol. 90, May 2020, Art. no. 106181.

[2] J.-W. Baek and K. Chung, "Context deep neural network model for predicting depression risk using multiple regression," *IEEE Access*, vol. 8, pp. 18171–18181, 2020.

[3] N. Sturm, A. Mayr, T. Le Van, V. Chupakhin, H. Ceulemans, J. Wegner, J.-F. Golib-Dzib, N. Jeliazkova, Y. Vandriessche, S. Böhm, V. Cima, J. Martinovic, N. Greene, T. V. Aa, T. J. Ashby, S. Hochreiter, O. Engkvist, G. Klambauer, and H. Chen, "Industry-scale application and evaluation of deep learning for drug target prediction," *J. Cheminformatics*, vol. 12, no. 1, p. 26, Dec. 2020.

[4] C. Chen, Y. Liu, X. Sun, C. D. Cairano-Gilfedder, and S. Titmus, "Automobile maintenance prediction using deep learning with GIS data," *Proc. CIRP*, vol. 81, pp. 447–452, Jan. 2019.

[5] J. Villalba-Diez, D. Schmidt, R. Gevers, J. Ordieres-Meré, M. Buchwitz, and W. Wellbrock, "Deep learning for industrial computer vision quality control in the printing Industry 4.0," *Sensors*, vol. 19, no. 18, p. 3987, Sep. 2019.

[6] A. Song, E. Seo, and H. Kim, "Anomaly VAE-Transformer: A deep learning approach for anomaly detection in decentralized finance," *IEEE Access*, vol. 11, pp. 98115–98131, 2023.

[7] M. Almutairi, L. A. Gabralla, S. Abubakar, and H. Chiroma, "Detecting elderly behaviors based on deep learning for healthcare: Recent advances, methods, real-world applications and challenges," *IEEE Access*, vol. 10, pp. 69802–69821, 2022.

[8] M. Li, W. Zhou, J. Liu, X. Zhang, F. Pan, H. Yang, M. Li, and D. Luo, "Vehicle interior noise prediction based on Elman neural network," *Appl. Sci.*, vol. 11, no. 17, p. 8029, Aug. 2021.

[9] H. B. Huang, J. H. Wu, X. R. Huang, M. L. Yang, and W. P. Ding, "The development of a deep neural network and its application to evaluating the interior sound quality of pure electric vehicles," *Mech. Syst. Signal Process.*, vol. 120, pp. 98–116, Apr. 2019.

[10] D. E. Tsokaktsidis, C. Nau, M. Maeder, and S. Marburg, "Using rectified linear unit and swish based artificial neural networks to describe noise transfer in a full vehicle context," *J. Acoust. Soc. Amer.*, vol. 150, no. 3, pp. 2088–2105, Sep. 2021.

[11] E. Koh, G. H. Nam, S. W. Kim, K. H. Park, and S. B. Kim, "Multi-sensor spectrogram transformer network for automobile noise prediction from electric power steering," *J. Korean Inst. Ind. Eng.*, vol. 48, no. 4, pp. 389–397, Aug. 2022.

[12] E. Ko, K. Jeong, Y. Jo, E. Koh, S. Ahn, G. H. Nam, S. W. Kim, K. H. Park, and S. B. Kim, "Noise level prediction from R-EPS automobile steering shaft sensors based on multi-encoder convolutional neural networks," *J. Korean Inst. Ind. Eng.*, vol. 48, no. 4, pp. 409–419, Aug. 2022.

[13] Y. Kang and J. Lee, "Randomized learning-based classification of sound quality using spectrogram image and time-series data: A practical perspective," *Eng. Appl. Artif. Intell.*, vol. 120, Apr. 2023, Art. no. 105867.

[14] J. Kim and J. Lee, "Statistical classification of vehicle interior sound through upsampling-based augmentation and correction using 1D CNN and LSTM," *IEEE Access*, vol. 10, pp. 100615–100626, 2022.

[15] L. Wyse, "Audio spectrogram representations for processing with convolutional neural networks," in *Proc. 1st Int. Conf. Deep Learn. Music*, 2017, pp. 37–41.

[16] D.-I. Noh, S.-G. Jeong, H.-T. Hoang, Q.-V. Pham, T. Huynh-The, M. Hasegawa, H. Sekiya, S.-Y. Kwon, S.-H. Chung, and W.-J. Hwang, "Signal preprocessing technique with noise-tolerant for RF-based UAV signal classification," *IEEE Access*, vol. 10, pp. 134785–134798, 2022.

[17] M. F. Aladdin, N. A. A. Jalil, N. Y. Guan, K. A. M. Rezali, and S. A. Adam, "Evaluation of human discomfort from combined noise and whole-body vibration in passenger vehicle," *Int. J. Automot. Mech. Eng.*, vol. 16, no. 2, pp. 6808–6824, Jul. 2019.

[18] A. Salazar, A. Rodríguez, N. Vargas, and L. Vergara, "On training road surface classifiers by data augmentation," *Appl. Sci.*, vol. 12, no. 7, p. 3423, Mar. 2022.

[19] D. Dablain, B. Krawczyk, and N. V. Chawla, "DeepSMOTE: Fusing deep learning and SMOTE for imbalanced data," *IEEE Trans. Neural Netw. Learn. Syst.*, 2023.

[20] X. Yang, Z. Song, I. King, and Z. Xu, "A survey on deep semi-supervised learning," *IEEE Trans. Knowl. Data Eng.*, 2022.

[21] W. Dai, X. Li, and K.-T. Cheng, "Semi-supervised deep regression with uncertainty consistency and variational model ensembling via Bayesian neural networks," in *Proc. AAAI Conf. Artif. Intell.*, 2023, pp. 7304–7313.

[22] Z.-H. Zhou and M. Li, "Semi-supervised regression with co-training," in *Proc. 19th Int. Joint Conf. Artif. Intell.*, 2005, pp. 908–913.

[23] S. J. Wetzel, R. G. Melko, and I. Tamblyn, "Twin neural network regression is a semi-supervised regression algorithm," *Mach. Learn., Sci. Technol.*, vol. 3, no. 4, Dec. 2022, Art. no. 045007.

[24] X. Jia, J. Yang, R. Liu, X. Wang, S. D. Cotofana, and W. Zhao, "Efficient computation reduction in Bayesian neural networks through feature decomposition and memorization," *IEEE Trans. Neural Netw. Learn. Syst.*, vol. 32, no. 4, pp. 1703–1712, Apr. 2021.

[25] Y. Du, K. Sekiguchi, Y. Bando, A. A. Nugraha, M. Fontaine, K. Yoshii, and T. Kawahara, "Semi-supervised multichannel speech separation based on a phone- and speaker-aware deep generative model of speech spectrograms," in *Proc. 28th Eur. Signal Process. Conf. (EUSIPCO)*, Jan. 2021, pp. 870–874.

[26] S. Wang, Z. Wu, G. He, S. Wang, H. Sun, and F. Fan, "Semi-supervised classification-aware cross-modal deep adversarial data augmentation," *Future Gener. Comput. Syst.*, vol. 125, pp. 194–205, Dec. 2021.

[27] S. Grollmisch and E. Cano, "Improving semi-supervised learning for audio classification with FixMatch," *Electronics*, vol. 10, no. 15, p. 1807, Jul. 2021.

[28] D. Berthelot, N. Carlini, I. Goodfellow, N. Papernot, A. Oliver, and C. A. Raffel, "MixMatch: A holistic approach to semi-supervised learning," in *Proc. Adv. Neural Inf. Process. Syst.*, 2019, pp. 5049–5059.

[29] D. Berthelot, N. Carlini, E. D. Cubuk, A. Kurakin, K. Sohn, H. Zhang, and C. Raffel, "ReMixMatch: Semi-supervised learning with distribution alignment and augmentation anchoring," in *Proc. Int. Conf. Learn. Represent.*, 2020, pp. 1–13.

[30] K. Sohn, D. Berthelot, N. Carlini, Z. Zhang, H. Zhang, C. A. Raffel, E. D. Cubuk, A. Kurakin, and C.-L. Li, "FixMatch: Implifying semi-supervised learning with consistency and confidence," in *Proc. Adv. Neural Inf. Process. Syst.*, vol. 33, 2020, pp. 596–608.

[31] B. Zhang, Y. Wang, W. Hou, H. A. O. Wu, J. Wang, M. Okumura, and T. Shinozaki, "FlexMatch: Boosting semi-supervised learning with curriculum pseudo labeling," in *Proc. Adv. Neural Inf. Process. Syst.*, vol. 34, 2021, pp. 18408–18419.

[32] M. Zheng, S. You, L. Huang, F. Wang, C. Qian, and C. Xu, "SimMatch: Semi-supervised learning with similarity matching," in *Proc. IEEE/CVF Conf. Comput. Vis. Pattern Recognit. (CVPR)*, Jun. 2022, pp. 14451–14461.
[33] N. Jean, S. M. Xie, and S. Ermon, "Semi-supervised deep kernel learning: Regression with unlabeled data by minimizing predictive variance," in *Proc. Adv. Neural Inf. Process. Syst.*, vol. 31, 2018, pp. 5327–5338.
[34] D. S. Park, W. Chan, Y. Zhang, C.-C. Chiu, B. Zoph, E. D. Cubuk, and Q. V. Le, "SpecAugment: A simple data augmentation method for automatic speech recognition," in *Proc. Interspeech*, 2019, pp. 2613–2617.
[35] H. Zhang, M. Cisse, Y. N. Dauphin, and D. Lopez-Paz, "mixup: Beyond empirical risk minimization," in *Proc. Int. Conf. Learn. Represent.*, 2018, pp. 1–13.
[36] S. Yao, A. Piao, W. Jiang, Y. Zhao, H. Shao, S. Liu, D. Liu, J. Li, T. Wang, S. Hu, L. Su, J. Han, and T. Abdelzaher, "STFNets: Learning sensing signals from the time-frequency perspective with short-time Fourier neural networks," in *Proc. World Wide Web Conf.*, May 2019, pp. 2192–2202.
[37] F. A. Bhatti, M. J. Khan, A. Selim, and F. Paisana, "Shared spectrum monitoring using deep learning," *IEEE Trans. Cogn. Commun. Netw.*, vol. 7, no. 4, pp. 1171–1185, Dec. 2021.
[38] D. Park, S. Lee, S. Park, and N. Kwak, "Radar-spectrogram-based UAV classification using convolutional neural networks," *Sensors*, vol. 21, no. 1, p. 210, Dec. 2020.
[39] C. Sandoval, M. N. Stolar, S. G. Hosking, D. Jia, and M. Lech, "Real-time team performance and workload prediction from voice communications," *IEEE Access*, vol. 10, pp. 78484–78492, 2022.

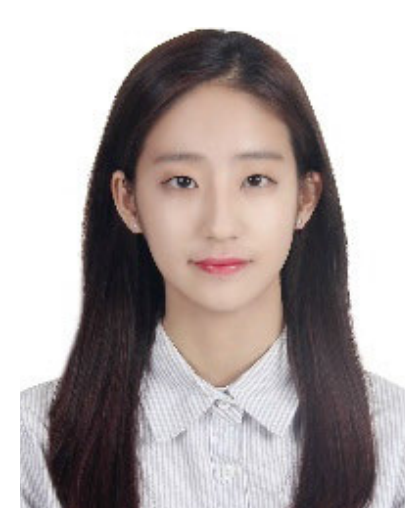

**SEJIN SIM** received the B.S. degree, in 2022. She is currently pursuing the Ph.D. degree with the School of Industrial and Management Engineering, Korea University, Seoul, South Korea. Her research interests include multivariate time series data analysis and semi-supervised learning.

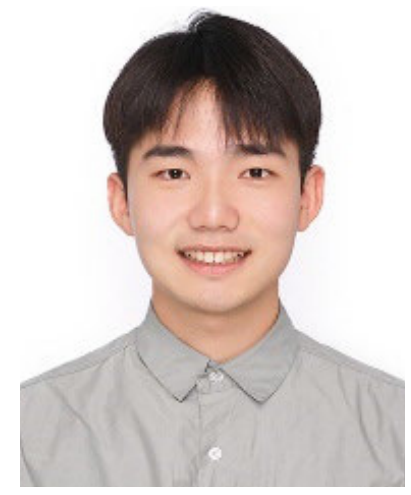

**JINSOO BAE** received the B.S. degree, in 2020. He is currently pursuing the Ph.D. degree with the School of Industrial and Management Engineering, Korea University, Seoul, South Korea. His research interests include uncertainty-aware deep neural networks and incomplete data analysis.

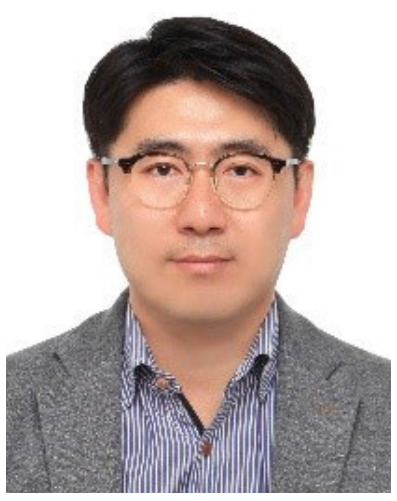

**SEOUNG BUM KIM** received the M.S. and Ph.D. degrees in industrial and systems engineering from the Georgia Institute of Technology, in 2001 and 2005, respectively. From 2005 to 2009, he was an Assistant Professor with the Department of Industrial and Manufacturing Systems Engineering, The University of Texas at Arlington. He is currently a Professor with the School of Industrial and Management Engineering, Korea University. He has published more than 180 internationally recognized journals and refereed conference proceedings. His research interest includes machine learning algorithms to create new methods for various problems appearing in engineering and science.

• • •